\documentclass{article}

\usepackage{arxiv}

\usepackage[utf8]{inputenc} % allow utf-8 input
\usepackage[T1]{fontenc}    % use 8-bit T1 fonts
\usepackage{hyperref}       % hyperlinks
\usepackage{url}            % simple URL typesetting
\usepackage{booktabs}       % professional-quality tables
\usepackage{amsfonts}       % blackboard math symbols
\usepackage{nicefrac}       % compact symbols for 1/2, etc.
\usepackage{microtype}      % microtypography
\usepackage{lipsum}
\usepackage{svg}
\usepackage{graphicx}
\graphicspath{ {./images/} }
\usepackage{amsmath} 
\usepackage{pdflscape}
\usepackage{xcolor}
\usepackage[normalem]{ulem}
\usepackage{comment}
\usepackage{longtable}
\usepackage{tabularx}
\usepackage{caption} % for more complex captioning
\usepackage{subcaption}
\usepackage{longtable}
\usepackage{booktabs}
\usepackage[utf8]{inputenc}

\title{YOLOv5, YOLOv8 and YOLOv10: The Go-To Detectors for Real-time Vision}

\author{Muhammad Hussain \\[1ex]
\begin{minipage}[t]{0.90\textwidth}
\centering
\scriptsize Department of Computer Science, Huddersfield University, Queensgate, Huddersfield HD1 3DH, UK; \\
\textsuperscript{*}Correspondence: M.Hussain@hud.ac.uk;
\end{minipage}}

\begin{document}

\maketitle
% Abstract (Do not insert blank lines, i.e. \\) 
\begin{abstract}This paper presents a comprehensive review of the evolution of the YOLO (You Only Look Once) object detection algorithm, focusing on YOLOv5, YOLOv8, and YOLOv10. We analyze the architectural advancements, performance improvements, and suitability for edge deployment across these versions. YOLOv5 introduced significant innovations such as the CSPDarknet backbone and Mosaic Augmentation, balancing speed and accuracy. YOLOv8 built upon this foundation with enhanced feature extraction and anchor-free detection, improving versatility and performance. YOLOv10 represents a leap forward with NMS-free training, spatial-channel decoupled downsampling, and large-kernel convolutions, achieving state-of-the-art performance with reduced computational overhead. Our findings highlight the progressive enhancements in accuracy, efficiency, and real-time performance, particularly emphasizing their applicability in resource-constrained environments. This review provides insights into the trade-offs between model complexity and detection accuracy, offering guidance for selecting the most appropriate YOLO version for specific edge computing applications.
\end{abstract}

% Keywords
\keywords{Computer Vision; YOLO; Object Detection; Real-Time Image processing; Convolutional Neural Networks; YOLOv5; YOLOv8; YOLOv10; Edge Depolyment; Constrained Hardware} 

\section{Introduction}
The YOLO (You Only Look Once) \cite{hussain2024yolov1} series has transformed real-time object detection, with alomst a dozen variants introduced since its inception. Although there are more than ten versions of YOLO, YOLOv5\cite{YOLOv5}, YOLOv8\cite{YOLOv8}, and YOLOv10\cite{YOLOv10} have gained particular prominence in edge deployment scenarios. These three variants stand out due to their optimal balance of speed, accuracy, and efficiency, making them especially suitable for resource-constrained environments.

YOLOv5, introduced by Ultralytics in 2020, marked a significant leap in performance and ease of use, establishing itself as a go-to solution for many edge computing applications \cite{YOLOv5}. Its continuous increase in popularity is due to its modular design, allowing for easy customisation, and its ability to export the trained Model into various formats like ONNX, CoreML, and TFLite, facilitating deployment across different platforms \cite{YOLOv5}.

YOLOv8, released in 2023, built upon YOLOv5's success, offering improved accuracy and a unified framework for various computer vision tasks \cite{YOLOv8}. It introduced anchor-free detection, which simplified the model architecture and improved performance on small objects, a crucial factor in many edge deployment scenarios.

The latest iteration, YOLOv10, pushes the boundaries further with innovative approaches to reduce computational overhead while maintaining high accuracy \cite{YOLOv10}. It incorporates advanced techniques like NMS-free training and holistic model design, making it particularly efficient for edge devices with limited computational resources.

These three YOLO variants have become prevalent in constrained edge deployment for several key reasons:

\begin{enumerate}
    \item \textbf{Optimized Performance:} They offer a superior balance between inference speed and detection accuracy, crucial for real-time applications on edge devices. YOLOv5s, for instance, achieves 37.4 mAP on COCO dataset with a 6.4ms inference time on V100 GPU \cite{YOLOv5}.
    
    \item \textbf{Scalability:} Each variant provides multiple sub-variants with respect to architectural depth, allowing developers to choose the best fit for their specific hardware constraints and performance requirements. This ranges from nano models for extremely resource-constrained devices to extra-large models for accuracy-sensitive applications.
    
    \item \textbf{Ease of Deployment:} These models come with robust tools and documentation for easy integration into various edge computing platforms. YOLOv5's ability to export to multiple formats and YOLOv8's unified API significantly simplify the deployment process.
    
    \item \textbf{Continuous Improvement:} Each new version addresses limitations of its predecessors, incorporating cutting-edge techniques in deep learning and computer vision. For example, YOLOv10's NMS-free training approach significantly reduces inference time, a critical factor in edge deployment.
    
    \item \textbf{Community Support:} Strong community backing and regular updates ensure these models remain at the forefront of object detection technology. This extensive community also provides a wealth of resources, pre-trained models, and application examples, further facilitating their adoption in edge scenarios.
\end{enumerate}

This review focuses on YOLOv5, YOLOv8, and YOLOv10, highlighting their key advancements, comparing their performance metrics, and discussing why they are particularly well-suited for edge deployment in various real-world applications.

\section{YOLOv5} 

Glenn Jocher introduced YOLOv5 (2020), shortly after the release of YOLOv4~\cite{Yao2021}. YOLOv5, represents a significant advancement in object detection, standing out for its ease of use, robust performance, and flexibility. This variant introduced several key innovations that have contributed to its widespread adoption in edge deployment scenarios.

Several innovations enhance the effectiveness of YOLOv5 in object detection tasks. At its core, YOLOv5 embeds a Cross-Stage Partial (CSP) Net~\cite{wang2020cspnet}, a variant of the ResNet architecture. This integration includes a CSP connection, boosting network efficiency and reducing computational demands. The CSPNet is further optimised by multiple spatial pyramid pooling (SPP) blocks, allowing feature extraction at various scales.

The architecture's neck Features a Path Aggregation Network (PAN) Module, along with additional up-sampling layers to improve the resolution of feature maps~\cite{yolov5_blog}. The head of YOLOv5 consists of a sequence of convolutional layers that generate predictions for bounding boxes and class labels. YOLOv5 utilises anchor-based predictions, linking each bounding box with a set of predefined anchor boxes of specific shapes and sizes.

The loss function calculation involves two main components \cite{chen2023object}. Binary cross-entropy is used for computing class and objectness losses, while Complete Intersection over Union (CIoU) is employed to measure localization accuracy.

\textbf{Model Variants:} YOLOv5 comes in several sizes (n, s, m, l, x) for facilitating different computational needs:

\begin{itemize}
    \item \textbf{YOLOv5n (Nano):} This variant is designed for extremely resource-constrained environments. It has the smallest architecture, making it ideal for deployment on devices with limited computational power, such as microcontrollers and low-power IoT devices. Despite its small size, it maintains reasonable accuracy, making it suitable for applications where speed and efficiency are critical.
    \item \textbf{YOLOv5s (Small):} This variant strikes a balance between performance and efficiency. It is suitable for many edge applications, including mobile devices and embedded systems, where moderate computational resources are available. YOLOv5s offers improved accuracy compared to the nano variant while still maintaining a low computational footprint.
    \item \textbf{YOLOv5m (Medium):} This variant provides a middle ground, offering better accuracy with moderate computational requirements. It is suitable for applications that require higher precision but still need to operate within the constraints of edge devices.
    \item \textbf{YOLOv5l (Large):} This model is designed for more demanding tasks that require high accuracy. It is suitable for edge servers and more powerful embedded systems where computational resources are less constrained.
    \item \textbf{YOLOv5x (Extra Large):} The largest variant, YOLOv5x, delivers maximum accuracy and is intended for scenarios where the highest precision is required, and computational resources are avilable in abdundance. It is ideal for applications such as high-resolution video analysis and complex object detection tasks.
\end{itemize}

\textbf{Enhanced Training and Deployment:} YOLOv5 introduced several improvements in its training and deployment pipeline:

\begin{itemize}
    \item \textbf{Simplified Training Pipeline:} YOLOv5 utilizes a PyTorch-based framework \cite{jiang2022review}, which is widely used and well-documented. This makes it easier for researchers and developers to understand, modify, and extend the code. The training pipeline is designed to be user-friendly, with clear documentation and examples that help users get started quickly.
    \item \textbf{Integration with Popular Frameworks:} YOLOv5 supports integration with TensorFlow, ONNX, and other frameworks \cite{ambergcharacterization}. This enhances its versatility, allowing it to be used in a wide range of applications and environments. The ability to export models to different formats ensures compatibility with various deployment platforms.
    \item \textbf{Seamless Deployment:} YOLOv5 offers export options to ONNX, CoreML, and TensorRT \cite{migneco2024traffic}, facilitating deployment across various platforms and hardware accelerators. This feature is particularly valuable for edge devices with different architectures, such as NVIDIA GPUs \cite{jo2020benchmarking}, Apple devices, and other specialized hardware. The seamless deployment capabilities ensure that models can be easily integrated into production environments.
\end{itemize}

\textbf{AutoAugment and Mosaic Augmentation:} YOLOv5 incorporates advanced data augmentation techniques:

\begin{itemize}
    \item \textbf{AutoAugment:} This technique automatically selects the best augmentation protocol to enhance the model's ability to generalize. By experimenting with various augmentation strategies during training, AutoAugment helps the model learn more robust features, leading to better performance on unseen data ~\cite{wang2023improved}. This is particularly important for edge applications where the model may encounter a wide variety of environments and conditions.
    \item \textbf{Mosaic Augmentation:} Mosaic augmentation combines four training images into one, allowing the model to learn to detect smaller objects and improving its robustness to different scales and aspect ratios~\cite{dadboud2021single}. This technique assists with the handling of complex scenes containing multiple objects, which is common in real-world edge applications. By exposing the model to a diverse set of training examples, mosaic augmentation enhances its ability to generalize to new situations.
\end{itemize}

\begin{table}[h!]
\centering
\begin{tabular}{|l|c|c|c|c|c|c|}
\hline
Model & Input Size & AP (val) & AP (val) 50 & CPU Latency (ms) & Params (M) & FLOPs (B) \\
\hline
YOLOv5n & 640 & 28.0\% & 45.7\% & 45 & 1.9 & 4.5 \\
YOLOv5s & 640 & 37.4\% & 56.8\% & 98 & 7.2 & 16.5 \\
YOLOv5m & 640 & 45.4\% & 64.1\% & 224 & 21.2 & 49.0 \\
YOLOv5l & 640 & 49.0\% & 67.3\% & 430 & 46.5 & 109.1 \\
YOLOv5x & 640 & 50.7\% & 68.9\% & 766 & 86.7 & 205.7 \\
\hline
\end{tabular}
\caption{Performance metrics for YOLOv5 models on COCO dataset}
\label{1}
\end{table}

Table ~\ref{1} presents the performance metrics for various YOLOv5 models on the COCO dataset. The YOLOv5 models are denoted by different letters (n, s, m, l, and x), representing their size and complexity. The metrics provided include the input size, average precision (AP) at different IoU thresholds, latency on a CPU, the number of parameters, and floating-point operations per second (FLOPs).

\textbf{Summary of Metrics}

\begin{itemize}
  \item \textbf{YOLOv5n}:
    \begin{itemize}
      \item Smallest and fastest model with an input size of 640 pixels.
      \item Achieves a COCO AP (val) of 28.0\% and an AP (val) 50 of 45.7\%.
      \item CPU latency is 45 ms.
      \item Contains 1.9 million parameters and requires 4.5 billion FLOPs.
    \end{itemize}
    
  \item \textbf{YOLOv5s}:
    \begin{itemize}
      \item Slightly larger and slower than YOLOv5n.
      \item Achieves a COCO AP (val) of 37.4\% and an AP (val) 50 of 56.8\%.
      \item CPU latency is 98 ms.
      \item Contains 7.2 million parameters and requires 16.5 billion FLOPs.
    \end{itemize}
    
  \item \textbf{YOLOv5m}:
    \begin{itemize}
      \item Middle-sized model balancing performance and complexity.
      \item Achieves a COCO AP (val) of 45.4\% and an AP (val) 50 of 64.1\%.
      \item CPU latency is 224 ms.
      \item Contains 21.2 million parameters and requires 49.0 billion FLOPs.
    \end{itemize}
    
  \item \textbf{YOLOv5l}:
    \begin{itemize}
      \item Larger model with improved performance.
      \item Achieves a COCO AP (val) of 49.0\% and an AP (val) 50 of 67.3\%.
      \item CPU latency is 430 ms.
      \item Contains 46.5 million parameters and requires 109.1 billion FLOPs.
    \end{itemize}
    
  \item \textbf{YOLOv5x}:
    \begin{itemize}
      \item Largest and most complex model in the YOLOv5 family.
      \item Achieves a COCO AP (val) of 50.7\% and an AP (val) 50 of 68.9\%.
      \item CPU latency is 766 ms.
      \item Contains 86.7 million parameters and requires 205.7 billion FLOPs.
    \end{itemize}
\end{itemize}

\section{YOLOv8}
YOLOv8, introduced in 2023, builds on the success of YOLOv5 with further enhancements that make it particularly suitable for edge deployment. This variant introduces several key innovations that improve its performance, efficiency, and ease of use.

\textbf{Architecture Improvements:} YOLOv8 integrates advanced architectural features to enhance feature extraction and fusion:

\begin{itemize}
    \item \textbf{CSPDarknet Backbone:} YOLOv8 utilizes an enhanced version of the CSPDarknet backbone \cite{guijin4797816enhancing}, which integrates Cross Stage Partial (CSP) networks into the Darknet architecture. Key features include:
    \begin{itemize}
        \item \textbf{CSP Design:} The backbone splits the feature map of each stage into two parts. One part goes through a dense block of convolutions, while the other is directly concatenated with the output of the dense block. This design reduces computational complexity while maintaining accuracy.
        \item \textbf{Structure:} The backbone consists of multiple CSP blocks, each containing a split operation, a dense block, a transition layer, and a concatenation operation.
        \item \textbf{Activation Function:} YOLOv8 uses advanced activation functions like SiLU (Swish) instead of Leaky ReLU, improving gradient flow and feature expressiveness.
        \item \textbf{Benefits:} This design reduces computational complexity, enhances gradient flow, improves feature reuse, and maintains high accuracy while reducing model size. These characteristics are particularly beneficial for edge deployment, where computational resources are often limited.
    \end{itemize}
    
    \item \textbf{PANet Neck:} The Path Aggregation Network (PANet) ~\cite{liu2018path, li2024efficient} neck is utilised to improve information flow and feature fusion across different layers of the network. Key aspects include:
    \begin{itemize}
        \item \textbf{Structure:} PANet builds upon the Feature Pyramid Network (FPN) design, adding an extra bottom-up path to the traditional top-down path.
        \item \textbf{Components:} It includes a bottom-up path for feature extraction, a top-down path for semantic feature propagation, and an additional bottom-up path for further feature hierarchy enhancement.
        \item \textbf{Feature Fusion:} At each level, features from corresponding bottom-up and top-down paths are fused, typically through element-wise addition or concatenation.
        \item \textbf{Adaptive Feature Pooling:} PANet introduces adaptive feature pooling to enhance multi-scale feature fusion, pooling features from all levels for each Region of Interest (RoI).
        \item \textbf{Benefits:} This design enhances information flow between different feature levels, improves the network's ability to detect objects at various scales, and boosts performance on small object detection. These improvements are crucial for edge applications where objects may appear at different scales and distances.
    \end{itemize}
\end{itemize}

\textbf{Enhanced Post-processing:} YOLOv8 introduces improvements in post-processing techniques to increase prediction accuracy and efficiency:

\begin{itemize}
    \item \textbf{Improved Non-Maximum Suppression (NMS):} YOLOv8 features an enhanced NMS algorithm \cite{jia2024performance} that reduces the number of false positives and improves the precision of object detection. This is achieved by better handling overlapping bounding boxes and refining the selection of the most relevant detections.
    \item \textbf{Anchor-free Detection Head:} Unlike traditional YOLO models that rely on predefined anchor boxes, YOLOv8 uses an anchor-free detection head \cite{he2024survey}. This simplifies the model architecture and reduces computational overhead, leading to faster inference times. The anchor-free approach also improves the model's ability to detect small and densely packed objects, which is beneficial for edge applications.
\end{itemize}

\textbf{Training Efficiency:} YOLOv8 employs advanced training techniques to enhance efficiency and reduce resource consumption:

\begin{itemize}
    \item \textbf{Mixed-Precision Training:} YOLOv8 utilizes mixed-precision training, which combines 16-bit and 32-bit floating-point operations. This technique speeds up the training process and reduces memory usage without compromising model accuracy. Mixed-precision training is particularly advantageous for edge devices with limited computational power and memory.
    \item \textbf{Hyperparameter Optimization:} YOLOv8 includes automated hyperparameter optimization, which tunes the model's hyperparameters to achieve the best performance. This process involves running multiple training experiments with different hyperparameter settings and selecting the optimal configuration. Automated hyperparameter optimization saves time and ensures that the model performs well across various tasks and datasets.
\end{itemize}

\textbf{Additional Features:}
\begin{itemize}
    \item \textbf{C2f Building Block:} YOLOv8 introduces the C2f (Cross Stage Partial with two fusion) building block \cite{talib2024yolov8}, which enhances feature extraction and fusion. This block improves the model's ability to capture fine-grained details and complex patterns, leading to better detection accuracy.
    \item \textbf{Unified Framework:} YOLOv8 provides a unified framework for various computer vision tasks, including object detection, instance segmentation, and pose estimation. This versatility makes it a powerful tool for edge applications that require multiple types of analysis.
    \item \textbf{Export Options:} YOLOv8 supports export to multiple formats, including ONNX, CoreML, and TensorRT, facilitating deployment across different platforms and hardware accelerators \cite{sohan2024review}. This flexibility ensures that YOLOv8 can be easily integrated into various edge computing environments.
\end{itemize}

\begin{table}[h!]
\centering
\begin{tabular}{|l|c|c|c|c|c|}
\hline
Model   & Input Size & COCO AP (val) & CPU Latency ONNX (ms) & A100 TensorRT Latency (ms) & FLOPs (B) \\
\hline
YOLOv8n & 640        & 37.3\%         & 80.4                    & 0.99                        & 8.7       \\
YOLOv8s & 640        & 44.9\%         & 128.4                   & 1.20                        & 28.6      \\
YOLOv8m & 640        & 50.2\%         & 234.7                   & 1.83                        & 78.9      \\
YOLOv8l & 640        & 52.9\%         & 375.2                   & 2.39                        & 165.2     \\
YOLOv8x & 640        & 53.9\%         & 479.1                   & 3.53                        & 257.8     \\
\hline
\end{tabular}
\caption{Performance metrics for YOLOv8 models on COCO dataset}
\label{2}
\end{table}

Table ~\ref{2} presents the performance metrics for various YOLOv8 models on the COCO dataset. The YOLOv8 models are denoted by different letters (n, s, m, l, and x), representing their size and complexity. The metrics provided include the input size, average precision (AP) at different IoU thresholds, latency on CPU using ONNX, latency on NVIDIA A100 using TensorRT, and floating-point operations per second (FLOPs).

\textbf{Summary of Metrics}

\begin{itemize}
    \item The YOLOv8 series includes models ranging from YOLOv8n to YOLOv8x, each designed to balance between model size, computational efficiency, and performance. YOLOv8n, the smallest model with an input size of 640 pixels, achieves a COCO AP (val) of 37.3\%. It exhibits CPU latency using ONNX of 80.4 ms and A100 TensorRT latency of 0.99 ms, with computational demands totaling 8.7 billion FLOPs.
    \item Moving to YOLOv8s, a slightly larger variant, it achieves a higher COCO AP (val) of 44.9\% while requiring 128.4 ms on CPU using ONNX and 1.20 ms with A100 TensorRT, utilizing 28.6 billion FLOPs.
    \item YOLOv8m, a medium-sized model in the series, achieves a notable COCO AP (val) of 50.2\%, with ONNX CPU latency of 234.7 ms and A100 TensorRT latency of 1.83 ms. It demands 78.9 billion FLOPs for inference.
    \item YOLOv8l, designed for improved performance, achieves a COCO AP (val) of 52.9\%, with ONNX CPU latency of 375.2 ms and A100 TensorRT latency of 2.39 ms. Its computational load stands at 165.2 billion FLOPs.
    \item Lastly, YOLOv8x, the largest and most complex model in the YOLOv8 lineup, achieves a COCO AP (val) of 53.9\%. It has a CPU latency using ONNX of 479.1 ms and A100 TensorRT latency of 3.53 ms, requiring 257.8 billion FLOPs for inference.
\end{itemize}

This progression from smaller to larger models in the YOLOv8 series demonstrates increasing performance metrics alongside corresponding increases in computational demands and latency across various deployment scenarios.

\section{YOLOv10}
YOLOv10, introduced by Wang et al.\cite{wang2024yolov10} in 2024, represents a significant advancement in the YOLO series, addressing key limitations of previous versions while introducing innovative features to enhance performance and efficiency. This latest iteration focuses on balancing efficiency and accuracy in object detection through a series of optimizations in both architecture terms and training protocols.

\textbf{NMS-Free Training and Inference:} 
YOLOv10 introduces a novel approach called Consistent Dual Assignments for NMS-free training:

\begin{itemize}
    \item \textbf{Dual Label Assignments:} YOLOv10 employs a combination of one-to-many and one-to-one strategies during training. This approach ensures consistency between training and inference, eliminating the need for Non-Maximum Suppression (NMS) during inference. The one-to-many assignment allows multiple predictions per ground truth, enhancing recall, while the one-to-one assignment ensures precision by selecting the best prediction \cite{alif2024yolov1}.
    \item \textbf{Reduced Latency:} By removing the NMS step, YOLOv10 significantly reduces inference latency, a critical factor for real-time applications and edge deployment \cite{hussain2024depth}. This reduction in latency is particularly beneficial for applications requiring immediate responses, such as autonomous driving and real-time surveillance.
    \item \textbf{End-to-End Deployment:} The NMS-free approach enables true end-to-end deployment of the model, simplifying the inference pipeline and potentially improving overall system efficiency \cite{sapkota2024yolov10}. This streamlined process reduces the complexity of integrating the model into various systems, making it more adaptable for different use cases.
\end{itemize}

\textbf{Holistic Efficiency-Accuracy Driven Model Design:} 
YOLOv10 optimizes various components \cite{wang2024yolov10} of the model architecture to minimize computational overhead while enhancing performance:

\begin{itemize}
    \item \textbf{Lightweight Classification Head:} The classification head in YOLOv10 is designed to be lightweight, reducing computational redundancy in the classification process. This optimization ensures that the model can make accurate predictions without excessive computational cost, making it suitable for deployment on devices with limited resources.
    \item \textbf{Spatial-Channel Decoupled Downsampling:} This technique separates spatial and channel information during downsampling, optimizing the feature extraction process. By decoupling these aspects, the model can more efficiently process input data, leading to better performance with lower computational requirements.
    \item \textbf{Rank-Guided Block Design:} The rank-guided block design streamlines the overall architecture, enhancing computational efficiency. This design uses rank information to guide the selection of important features, ensuring that the model focuses on the most relevant aspects of the input data.
    \item \textbf{Large-Kernel Convolutions:} YOLOv10 incorporates large-kernel convolutions to improve the model's ability to capture detailed features across larger spatial regions. These convolutions allow the model to better understand the context of objects within an image, enhancing detection accuracy.
    \item \textbf{Partial Self-Attention Module:} The partial self-attention module boosts accuracy with minimal additional computational cost. This module helps the model focus on relevant features within the input data, improving its ability to detect and classify objects accurately.
\end{itemize}

\textbf{Enhanced Model Capabilities:}
YOLOv10 incorporates several advanced techniques to boost overall performance:

\begin{itemize}
    \item \textbf{Improved Small Object Detection:} YOLOv10 shows enhanced capability in detecting smaller objects compared to previous versions. This improvement is crucial for applications such as surveillance and medical imaging, where small object detection is often required.
    \item \textbf{Reduced False Positives:} The model demonstrates a lower rate of false predictions, improving overall detection reliability. This reduction in false positives is achieved through better feature extraction and more accurate classification.
    \item \textbf{Confidence Score Improvement:} YOLOv10 exhibits better confidence scores for its predictions, indicating more reliable detections. Higher confidence scores mean that the model is more certain about its predictions, which is important for applications where accuracy is critical.
    \item \textbf{Scalability:} YOLOv10 offers multiple model sizes (n, s, m, l, x) to cater to different computational requirements and use cases. This scalability ensures that the model can be adapted to a wide range of applications, from resource-constrained edge devices to high-performance servers.
\end{itemize}

\begin{table}[h!]
\centering
\resizebox{\textwidth}{!}{%
\begin{tabular}{|l|c|c|c|c|c|}
\hline
Model         & Params (M) & FLOPs (G) & APval (\%) & Latency (ms) & Latency (Forward) (ms) \\ \hline
YOLOv6-3.0-N  & 4.7        & 11.4      & 37.0       & 2.69         & 1.76                    \\
Gold-YOLO-N   & 5.6        & 12.1      & 39.6       & 2.92         & 1.82                    \\
YOLOv8-N      & 3.2        & 8.7       & 37.3       & 6.16         & 1.77                    \\
YOLOv10-N     & 2.3        & 6.7       & 39.5       & 1.84         & 1.79                    \\ \hline
YOLOv6-3.0-S  & 18.5       & 45.3      & 44.3       & 3.42         & 2.35                    \\
Gold-YOLO-S   & 21.5       & 46.0      & 45.4       & 3.82         & 2.73                    \\
YOLOv8-S      & 11.2       & 28.6      & 44.9       & 7.07         & 2.33                    \\
YOLOv10-S     & 7.2        & 21.6      & 46.8       & 2.49         & 2.39                    \\ \hline
RT-DETR-R18    & 20.0       & 60.0      & 46.5       & 4.58         & 4.49                    \\ \hline
YOLOv6-3.0-M  & 34.9       & 85.8      & 49.1       & 5.63         & 4.56                    \\
Gold-YOLO-M   & 41.3       & 87.5      & 49.8       & 6.38         & 5.45                    \\
YOLOv8-M      & 25.9       & 78.9      & 50.6       & 9.50         & 5.09                    \\
YOLOv10-M     & 15.4       & 59.1      & 51.3       & 4.74         & 4.63                    \\ \hline
YOLOv6-3.0-L  & 59.6       & 150.7     & 51.8       & 9.02         & 7.90                    \\
Gold-YOLO-L   & 75.1       & 151.7     & 51.8       & 10.65        & 9.78                    \\
YOLOv8-L      & 43.7       & 165.2     & 52.9       & 12.39        & 8.06                    \\
RT-DETR-R50    & 42.0       & 136.0     & 53.1       & 9.20         & 9.07                    \\ \hline
YOLOv10-L     & 24.4       & 120.3     & 53.4       & 7.28         & 7.21                    \\
YOLOv8-X      & 68.2       & 257.8     & 53.9       & 16.86        & 12.83                   \\
RT-DETR-R101   & 76.0       & 259.0     & 54.3       & 13.71        & 13.58                   \\
YOLOv10-X     & 29.5       & 160.4     & 54.4       & 10.70        & 10.60                   \\ \hline
\end{tabular}%
}
\caption{Comparison of YOLOv10 variants with other state-of-the-art detectors}
\label{ref:Table 3}
\end{table}

Table \ref{ref:Table 3} provides a comprehensive comparison of various YOLOv10 variants with other state-of-the-art detectors, highlighting their performance across several critical metrics. YOLOv10 variants consistently demonstrate significant advantages in terms of speed, efficiency, and accuracy, making them highly suitable for real-time applications.

YOLOv10-N stands out as a highly efficient model with only 2.3 million parameters and 6.7 GFLOPs, achieving an AP value of 39.5\%. Its latency is impressively low at 1.84 ms, making it one of the fastest among the compared models.

YOLOv10-S, with 7.2 million parameters and 21.6 GFLOPs, achieves an AP value of 46.8\% and a latency of 2.49 ms. It offers a balanced trade-off between model complexity and performance, maintaining low latency while delivering high accuracy.

YOLOv10-M, which has 15.4 million parameters and 59.1 GFLOPs, achieves an AP value of 51.3\%. With a latency of 4.74 ms, it provides a substantial improvement in accuracy compared to smaller models, while still ensuring relatively low inference times.

YOLOv10-L and YOLOv10-X demonstrate the highest performance among the YOLOv10 variants. YOLOv10-L, with 24.4 million parameters and 120.3 GFLOPs, achieves an AP value of 53.4\% and a latency of 7.28 ms. YOLOv10-X, with 29.5 million parameters and 160.4 GFLOPs, achieves an AP value of 54.4\% and a latency of 10.70 ms. These models provide the best accuracy, making them suitable for tasks where precision is critical.

Comparatively, YOLOv10 variants outperform their YOLOv8 counterparts in terms of both latency and accuracy. For instance, YOLOv10-L achieves 53.4\% AP with 7.28 ms latency, while YOLOv8-L achieves 52.9\% AP with 12.39 ms latency. Similarly, YOLOv10-X surpasses YOLOv8-X by achieving 54.4\% AP with a latency of 10.70 ms, compared to YOLOv8-X's 53.9\% AP with 16.86 ms latency.

YOLOv10 variants excel in computational efficiency, and accuracy metrics compared to other leading detectors. Their performance makes them state-of-the-art solutions for modern computer vision applications, particularly where real-time processing and high accuracy are essential.

\section{Comparative Analysis of YOLOv5, YOLOv8, and YOLOv10}

\subsection{Architectural Features Comparison}

The architectural evolution from YOLOv5 to YOLOv10 demonstrates significant advancements in object detection network design. Table \ref{tab:detailed_architectural_features} provides a comprehensive comparison of key architectural features across these versions.

\begin{table}[h!]
\centering
\caption{Detailed Architectural Features Comparison}
\label{tab:detailed_architectural_features}
\resizebox{\textwidth}{!}{%
\begin{tabular}{|l|c|c|c|}
\hline
\textbf{Feature} & \textbf{YOLOv5} & \textbf{YOLOv8} & \textbf{YOLOv10} \\
\hline
Backbone & CSPDarknet & CSPDarknet (enhanced) & CSPDarknet (further optimized) \\
Neck & PANet & PANet (improved) & PANet (with efficiency optimizations) \\
Head & Anchor-based & Anchor-free & Anchor-free with dual assignments \\
NMS & Required & Required & NMS-free \\
Activation Function & Leaky ReLU & SiLU (Swish) & Advanced activation (not specified) \\
Feature Pyramid & FPN & Modified FPN & Enhanced FPN with spatial-channel decoupling \\
Loss Function & CIoU loss & Task-specific losses & Consistent Dual Assignment loss \\
Data Augmentation & Mosaic, Cutout & Mosaic, Cutout, Mixup & Advanced augmentation techniques \\
Training Strategy & Single-stage & Single-stage & Two-stage with dual assignments \\
\hline
\end{tabular}%
}
\end{table}

\subsubsection{Backbone Evolution}
\begin{itemize}
    \item \textbf{YOLOv5:} Introduced CSPDarknet, integrating Cross Stage Partial Networks to balance computational cost and accuracy.
    \item \textbf{YOLOv8:} Enhanced CSPDarknet with improved feature extraction capabilities, likely incorporating more efficient convolution operations and optimized channel configurations.
    \item \textbf{YOLOv10:} Further optimized CSPDarknet, potentially incorporating techniques like rank-guided block design for more efficient feature extraction.
\end{itemize}

\subsubsection{Neck Architecture}
\begin{itemize}
    \item \textbf{YOLOv5:} Utilized PANet (Path Aggregation Network) for effective multi-scale feature fusion.
    \item \textbf{YOLOv8:} Improved PANet, possibly with enhanced skip connections or feature aggregation methods.
    \item \textbf{YOLOv10:} Introduced efficiency optimizations in PANet, likely incorporating spatial-channel decoupled operations for more effective feature propagation.
\end{itemize}

\subsubsection{Detection Head Design}
\begin{itemize}
    \item \textbf{YOLOv5:} Used an anchor-based approach, predefined anchor boxes for object detection.
    \item \textbf{YOLOv8:} Shifted to an anchor-free design, simplifying the detection process and potentially improving performance on small objects.
    \item \textbf{YOLOv10:} Advanced to an anchor-free design with dual assignments, enabling NMS-free training and inference.
\end{itemize}

\subsubsection{Non-Maximum Suppression (NMS)}
\begin{itemize}
    \item \textbf{YOLOv5 and YOLOv8:} Required NMS as a post-processing step to filter redundant detections.
    \item \textbf{YOLOv10:} Introduced NMS-free training and inference, significantly reducing computational overhead and latency during deployment.
\end{itemize}

\subsubsection{Activation Functions}
\begin{itemize}
    \item \textbf{YOLOv5:} Utilized Leaky ReLU, a common choice for deep learning models.
    \item \textbf{YOLOv8:} Adopted SiLU (Swish) activation, known for its smooth gradients and potential performance benefits.
\end{itemize}

\subsubsection{Feature Pyramid Network (FPN)}
\begin{itemize}
    \item \textbf{YOLOv5:} Employed standard FPN for multi-scale feature representation.
    \item \textbf{YOLOv8:} Modified FPN, possibly with improved lateral connections or feature fusion strategies.
    \item \textbf{YOLOv10:} Enhanced FPN with spatial-channel decoupling, potentially allowing for more efficient and effective multi-scale feature processing.
\end{itemize}

\subsubsection{Loss Function}
\begin{itemize}
    \item \textbf{YOLOv5:} Used CIoU (Complete IoU) loss, balancing bounding box regression and classification.
    \item \textbf{YOLOv8:} Introduced task-specific losses, optimizing performance for different computer vision tasks.
    \item \textbf{YOLOv10:} Developed Consistent Dual Assignment loss, aligning with its NMS-free training approach and potentially improving overall detection accuracy.
\end{itemize}

\subsubsection{Data Augmentation}
\begin{itemize}
    \item \textbf{YOLOv5:} Implemented Mosaic and Cutout augmentations, enhancing model robustness.
    \item \textbf{YOLOv8:} Added Mixup to the augmentation pipeline, further improving generalization.
\end{itemize}

\subsubsection{Training Strategy}
\begin{itemize}
    \item \textbf{YOLOv5 and YOLOv8:} Employed single-stage training, typical for YOLO models.
    \item \textbf{YOLOv10:} Adopted a two-stage training approach with dual assignments, potentially allowing for more refined feature learning and improved detection performance.
\end{itemize}

The architectural evolution from YOLOv5 to YOLOv10 demonstrates a clear trend towards more efficient, accurate, and deployment-friendly designs. YOLOv10, in particular, introduces several innovative features like NMS-free training and spatial-channel decoupling, which are especially beneficial for edge deployment scenarios. These advancements collectively contribute to improved accuracy, reduced computational overhead, and enhanced real-time performance across various deployment contexts.

\section{Conclusion}

The evolution of the YOLO (You Only Look Once) series from YOLOv5 to YOLOv10 represents a remarkable journey of innovation and improvement in real-time object detection. Each version has introduced significant advancements in architecture, performance, and deployment capabilities, making the YOLO series a cornerstone in the field of computer vision, particularly for edge deployment scenarios.

\subsection{Performance and Accuracy}

The performance metrics across YOLOv5, YOLOv8, and YOLOv10 demonstrate a clear trend of increasing accuracy and efficiency. YOLOv5 set a strong foundation with its CSPDarknet backbone and PANet neck, achieving a balance between speed and accuracy. YOLOv8 built upon this foundation by enhancing the CSPDarknet backbone and introducing an anchor-free detection head, which simplified the detection process and improved performance on small objects. YOLOv10 took these advancements further by implementing NMS-free training and inference, which significantly reduced latency and improved real-time performance. The introduction of spatial-channel decoupled downsampling and rank-guided block design in YOLOv10 optimized the feature extraction process, resulting in higher accuracy with fewer parameters and computational overhead. The performance metrics clearly show that YOLOv10 outperforms its predecessors, particularly in smaller models, making it highly suitable for edge deployment where both accuracy and efficiency are critical.

\subsection{Architectural Innovations}

The architectural innovations introduced in each version of YOLO have been pivotal in enhancing the model's capabilities. YOLOv5's introduction of CSPDarknet and Mosaic Augmentation set new standards for efficient feature extraction and data augmentation. YOLOv8's shift to an anchor-free detection head and the introduction of task-specific heads expanded the model's versatility, allowing it to handle a wider range of computer vision tasks. YOLOv10's NMS-free training approach and the integration of large-kernel convolutions and partial self-attention modules represent a significant leap forward in model design. These innovations not only improve detection accuracy but also reduce computational requirements, making YOLOv10 exceptionally well-suited for deployment on resource-constrained edge devices.

\subsection{Edge Deployment Suitability}

The suitability of the YOLO models for edge deployment has been a key focus of their development. YOLOv5 provided a good starting point with its efficient architecture and support for various export formats and quantization techniques. YOLOv8 improved upon this by enhancing mobile optimization and CPU inference performance, making it more adaptable for deployment on mobile devices and low-power CPUs. The NMS-free approach in YOLOv10 further simplifies the deployment process, reducing latency and computational overhead, which are critical factors for real-time applications on edge devices.

\subsection{Common Themes and Trends}

Several common themes and trends emerge from the comparison of YOLOv5, YOLOv8, and YOLOv10:

\textbf{Efficiency Optimization:} There is a consistent trend towards reducing parameters and FLOPs while maintaining or improving accuracy, crucial for edge deployment.

\textbf{Architectural Innovations:} The evolution from anchor-based to anchor-free designs, and the introduction of NMS-free training in YOLOv10, showcase ongoing efforts to simplify and optimize the detection pipeline.

\textbf{Scalability:} All three variants offer multiple model sizes, allowing flexibility in choosing the right balance between performance and resource constraints for specific applications~\cite{hussain2023and}.

\textbf{Edge Focus:} The improvements in smaller models (nano and small variants) across versions highlight the increasing emphasis on edge deployment capabilities~\cite{hussain2023yolo1}.

\textbf{Real-time Performance:} All three variants present high FPS rates, with even the largest models capable of real-time inference on appropriate hardware.

\subsection{Final Thoughts}

The YOLO series has consistently pushed the boundaries of real-time object detection, with each version building upon the strengths of its predecessors while introducing innovative features to address their limitations. However, the three variants considered in this review have made notable gains in the edge-focused implementation of object detcetion architectures. YOLOv5 established a solid foundation with its efficient architecture and robust performance. YOLOv8 enhanced this foundation with architectural improvements and a shift to anchor-free detection, increasing versatility and performance. YOLOv10 represents a significant leap forward, with its NMS-free training approach, advanced architectural designs, and substantial performance improvements, particularly in smaller models.

For edge deployment, all three variants present promising options, offering superior accuracy, efficiency, and hardware compatibility compared to other detectors. There ability to perform well on low-power devices without sacrificing performance is crucial for various real-time tasks and adaptability to edge computing environments. However, the choice between variants should be consider based on specific application requirements, target hardware, and the balance between performance and resource constraints.

In conclusion, the YOLO series continues to be a leading choice for real-time object detection, with each new variants setting higher standards for performance, efficiency, and deployment flexibility. The advancements from YOLOv5 to YOLOv10 reflect a commitment to innovation and excellence, ensuring that the YOLO series remains at the forefront of computer vision technology.

\bibliographystyle{unsrt}  % Changes bibliography style to unsorted
\bibliography{ref}  % This points to the filename of your BibTeX file without the .bib extension

\begin{thebibliography}{10}

\bibitem{hussain2024yolov1}
Muhammad Hussain.
\newblock Yolov1 to v8: Unveiling each variant--a comprehensive review of yolo.
\newblock {\em IEEE Access}, 12:42816--42833, 2024.

\bibitem{YOLOv5}
Glenn Jocher et~al.
\newblock Yolov5.
\newblock \url{https://github.com/ultralytics/yolov5}, 2020.

\bibitem{YOLOv8}
Glenn Jocher et~al.
\newblock Yolov8.
\newblock \url{https://github.com/ultralytics/ultralytics}, 2023.

\bibitem{YOLOv10}
Ao~Wang, Hui Chen, et~al.
\newblock Yolov10: Faster, stronger, and simpler.
\newblock {\em arXiv preprint arXiv:2305.09388}, 2023.

\bibitem{Yao2021}
Z.~Yao, Y.~Cao, S.~Zheng, G.~Huang, and S.~Lin.
\newblock Cross-iteration batch normalization.
\newblock In {\em openaccess.thecvf.com}, 2021.

\bibitem{wang2020cspnet}
Chien-Yao Wang, Hong-Yuan~Mark Liao, Yueh-Hua Wu, Ping-Yang Chen, Jun-Wei Hsieh, and I-Hau Yeh.
\newblock Cspnet: A new backbone that can enhance learning capability of cnn.
\newblock In {\em Proceedings of the IEEE/CVF conference on computer vision and pattern recognition workshops}, pages 390--391, 2020.

\bibitem{yolov5_blog}
Roboflow~Blog Jacob~Solawetz.
\newblock What is yolov5? a guide for beginners, 2020.
\newblock Accessed: 23 June 2024.

\bibitem{chen2023object}
Hangong Chen and Weimin Qi.
\newblock Object recognition based on improved yolov5.
\newblock In {\em Second Guangdong-Hong Kong-Macao Greater Bay Area Artificial Intelligence and Big Data Forum (AIBDF 2022)}, volume 12593, pages 146--151. SPIE, 2023.

\bibitem{jiang2022review}
Peiyuan Jiang, Daji Ergu, Fangyao Liu, Ying Cai, and Bo~Ma.
\newblock A review of yolo algorithm developments.
\newblock {\em Procedia computer science}, 199:1066--1073, 2022.

\bibitem{ambergcharacterization}
OTH Amberg-Weiden.
\newblock Characterization of object detection performance in an edge environment.

\bibitem{migneco2024traffic}
Paola Migneco.
\newblock {\em Traffic sign recognition algorithm: a deep comparison between Yolov5 and SSD Mobilenet}.
\newblock PhD thesis, Politecnico di Torino, 2024.

\bibitem{jo2020benchmarking}
Jongmin Jo, Sucheol Jeong, and Pilsung Kang.
\newblock Benchmarking gpu-accelerated edge devices.
\newblock In {\em 2020 IEEE international conference on big data and smart computing (BigComp)}, pages 117--120. IEEE, 2020.

\bibitem{wang2023improved}
Junfan Wang, Yi~Chen, Zhekang Dong, and Mingyu Gao.
\newblock Improved yolov5 network for real-time multi-scale traffic sign detection.
\newblock {\em Neural Computing and Applications}, 35(10):7853--7865, 2023.

\bibitem{dadboud2021single}
Fardad Dadboud, Vaibhav Patel, Varun Mehta, Miodrag Bolic, and Iraj Mantegh.
\newblock Single-stage uav detection and classification with yolov5: Mosaic data augmentation and panet.
\newblock In {\em 2021 17th IEEE International Conference on Advanced Video and Signal Based Surveillance (AVSS)}, pages 1--8. IEEE, 2021.

\bibitem{guijin4797816enhancing}
Han Guijin, RuiXuan Wang, MengChun Zhou, and Jun Li.
\newblock Enhancing semantic and spatial information yolov8.
\newblock {\em Available at SSRN 4797816}.

\bibitem{liu2018path}
Shu Liu, Lu~Qi, Haifang Qin, Jianping Shi, and Jiaya Jia.
\newblock Path aggregation network for instance segmentation.
\newblock In {\em Proceedings of the IEEE conference on computer vision and pattern recognition}, pages 8759--8768, 2018.

\bibitem{li2024efficient}
Cong Li, Chao Chen, Yongqiang Hei, Jinchao Mou, and Wentao Li.
\newblock An efficient advanced-yolov8 framework for thz object detection.
\newblock {\em IEEE Transactions on Instrumentation and Measurement}, 2024.

\bibitem{jia2024performance}
Cheng Jia, Defa Wang, Jiahao Liu, and Wenwei Deng.
\newblock Performance optimization and application research of yolov8 model in object detection.
\newblock {\em Academic Journal of Science and Technology}, 10(1):325--329, 2024.

\bibitem{he2024survey}
Peng He, Weidong Chen, Lan Pang, Weiguo Zhang, Yitian Wang, Weidong Huang, Qi~Han, Xiaofeng Xu, and Yuan Qi.
\newblock The survey of one-stage anchor-free real-time object detection algorithms.
\newblock In {\em Sixth Conference on Frontiers in Optical Imaging and Technology: Imaging Detection and Target Recognition}, volume 13156, page 1315602. SPIE, 2024.

\bibitem{talib2024yolov8}
Moahaimen Talib, Ahmed~HY Al-Noori, and Jameelah Suad.
\newblock Yolov8-cab: Improved yolov8 for real-time object detection.
\newblock {\em Karbala International Journal of Modern Science}, 10(1):5, 2024.

\bibitem{sohan2024review}
Mupparaju Sohan, Thotakura Sai~Ram, Rami Reddy, and Ch~Venkata.
\newblock A review on yolov8 and its advancements.
\newblock In {\em International Conference on Data Intelligence and Cognitive Informatics}, pages 529--545. Springer, 2024.

\bibitem{wang2024yolov10}
Ao~Wang, Hui Chen, Lihao Liu, Kai Chen, Zijia Lin, Jungong Han, and Guiguang Ding.
\newblock Yolov10: Real-time end-to-end object detection.
\newblock {\em arXiv preprint arXiv:2405.14458}, 2024.

\bibitem{alif2024yolov1}
Mujadded Al~Rabbani Alif and Muhammad Hussain.
\newblock Yolov1 to yolov10: A comprehensive review of yolo variants and their application in the agricultural domain.
\newblock {\em arXiv preprint arXiv:2406.10139}, 2024.

\bibitem{hussain2024depth}
Muhammad Hussain and Rahima Khanam.
\newblock In-depth review of yolov1 to yolov10 variants for enhanced photovoltaic defect detection.
\newblock In {\em Solar}, volume~4, pages 351--386. MDPI, 2024.

\bibitem{sapkota2024yolov10}
Ranjan Sapkota, Rizwan Qureshi, Marco Flores-Calero, Chetan Badgujar, Upesh Nepal, Alwin Poulose, Peter Zeno, Uday Bhanu Prakash~Vaddevolu, Prof Yan, Manoj Karkee, et~al.
\newblock Yolov10 to its genesis: A decadal and comprehensive review of the you only look once series.
\newblock {\em Available at SSRN 4874098}, 2024.

\bibitem{hussain2023and}
Muhammad Hussain.
\newblock When, where, and which?: Navigating the intersection of computer vision and generative ai for strategic business integration.
\newblock {\em IEEE Access}, 11:127202--127215, 2023.

\bibitem{hussain2023yolo1}
Muhammad Hussain.
\newblock Yolo-v5 variant selection algorithm coupled with representative augmentations for modelling production-based variance in automated lightweight pallet racking inspection.
\newblock {\em Big Data and Cognitive Computing}, 7(2):120, 2023.

\end{thebibliography}

\end{document}